\documentclass[9pt]{article}
\usepackage{spconf,amsmath,graphicx}
\usepackage{ifpdf}
\pdfoutput=1
\usepackage{tightenum}
\usepackage{comment}
\usepackage[utf8]{inputenc}
\usepackage[scaled=0.8]{beramono}
\usepackage[T1]{fontenc}      

\usepackage{float}
\usepackage{listings}
\usepackage{amsmath}
\usepackage{color}

\usepackage{mathtools}
\usepackage[english]{babel}
\usepackage{graphicx}
\usepackage{url}
\usepackage{tikz}
\usepackage{setspace}
\newif\ifremark
\long\def\remark#1{
\ifremark%
        \begingroup%
        \dimen0=\columnwidth
        \advance\dimen0 by -1in%
        \setbox0=\hbox{\parbox[b]{\dimen0}{\protect\em #1}}
        \dimen1=\ht0\advance\dimen1 by 2pt%
        \dimen2=\dp0\advance\dimen2 by 2pt%
        \vskip 0.25pt%
        \hbox to \columnwidth{%
                \vrule height\dimen1 width 3pt depth\dimen2%
                \hss\copy0\hss%
                \vrule height\dimen1 width 3pt depth\dimen2%
        }%
        \endgroup%
\fi}

\newcommand{\dnns}{\textmd{\textsc{dnn}s}}
\newcommand{\dlc}{\textmd{\textsc{dLAC}}}

\newcommand{\figtitle}[1]{\textbf{#1}}

\newcommand{\fixme}[1]{#1}

\newcommand{\qasicBold}[1]{\textmd{\textsc{\textbf{QsCore}}}}
\newcommand{\qasicsBold}[1]{\textmd{\textsc{\textbf{QsCores}}}}

\remarktrue

\renewcommand{\em}{\it}

\newcommand{\ignore}[1]{}

\newcommand{\mypar}[1]{\vspace{.1in}\noindent{\textbf {#1}} \hspace{.04in}}
\newcommand{\boldparagraph}[1]{\vspace*{-1ex}\mypar{#1}}

\def\dwfigure[#1,#2,#3,#4]{

\begin{figure*}[h]
\begin{minipage}[c]{\textwidth}{
\vspace{0mm}
\begin{center}
\includegraphics[width=\textwidth]{#1}
\vspace{-4mm}
\includegraphics[width=\textwidth]{#2}
\vspace{-4mm}
\caption[]{#3} 
\label{#4}
 
\end{center}

}\end{minipage}
\end{figure*}
}

\def\twfigure[#1,#2,#3,#4,#5]{

\begin{figure}[H]
\begin{minipage}[c]{\textwidth}{
\vspace{0mm}
\begin{center}
\includegraphics[width=\textwidth]{#1}
\vspace{-4mm}
\includegraphics[width=\textwidth]{#2}
\vspace{-4mm}
\includegraphics[width=\textwidth]{#3}
\vspace{-4mm}
\caption[]{#4} 
\label{#5}
 
\end{center}
}\end{minipage}
\end{figure}
}

\def\cfigurelong[#1,#2,#3]{
\begin{figure}
\vspace*{-3mm}
\begin{center}

\includegraphics[height=0.5\textheight]{#1} 
 
\vspace*{-6mm}
\caption[]{#2
} \label{#3}
 
\vspace*{0mm}
\end{center}
\vspace*{0mm}
\end{figure}}

\def\wcfigurelong[#1,#2,#3]{
\begin{figure}
\vspace*{-3mm}
\begin{center}

\includegraphics[height=0.77\textheight]{#1} 
 
\vspace*{0mm}
\caption[]{#2
} \label{#3}
 
\vspace*{0mm}
\end{center}
\vspace*{0mm}
\end{figure}}

\def\cfigureslim[#1,#2,#3]{
\begin{figure}
\vspace*{0mm}
\begin{center}

\includegraphics[width=0.8\columnwidth]{#1} 
 
\vspace*{0mm}
\caption[]{#2
} \label{#3}
 
\vspace*{0mm}
\end{center}
\vspace*{0mm}
\end{figure}}

\def\cfigure[#1,#2,#3]{
\begin{figure}
\vspace*{-2mm}
\begin{center}

\includegraphics[width=\columnwidth]{#1} 
 
\vspace*{-4mm}
\caption[]{#2
} \label{#3}
 
\vspace*{-3mm}
\end{center}
\vspace*{-3mm}
\end{figure}}

\def\cfigurebigger[#1,#2,#3]{
\begin{figure}
\begin{center}

\includegraphics[clip=true, trim=0in 0in 0.35in 0in, width=\columnwidth]{#1} 
 
\caption[]{#2
} \label{#3}
 
\vspace*{0mm}
\end{center}
\vspace*{0mm}
\end{figure}}

\def\cfiguretemp[#1,#2,#3]{
\begin{figure}
\vspace*{0mm}
\begin{center}

\includegraphics[width=\textwidth]{#1} 
 
\vspace*{0mm}\caption[]{#2
} \label{#3}
 
\vspace*{0mm}
\end{center}
\vspace*{0mm}
\end{figure}}

\def\wfigure[#1,#2,#3]{
\begin{figure*}
\vspace*{0mm}
\begin{center}
 
\includegraphics[width=0.8\textwidth]{#1} 
 
\vspace*{0mm}\caption[]{#2
} \label{#3}
 
\vspace*{0mm}
\end{center}

\end{figure*}}

\def\wfigureslim[#1,#2,#3]{
\begin{figure*}
\vspace*{0mm}
\begin{center}
 
\includegraphics[width=0.7\textwidth]{#1} 
 
\vspace*{0mm}\caption[]{#2
} \label{#3}
 
\vspace*{0mm}
\end{center}

\end{figure*}}

\def\wfigurewide[#1,#2,#3]{
\begin{figure*}
\vspace*{0mm}
\begin{center}
 
\includegraphics[width=0.92\textwidth]{#1} 
 
\vspace*{0mm}\caption[]{#2
} \label{#3}
 
\vspace*{0mm}
\end{center}
\vspace*{-2mm}
\end{figure*}}

\def\threefigure[#1,#2,#3,#4,#5]{
\begin{figure*}
\vspace*{0mm}
\begin{center}

\begin{tabular}{ccc}
\includegraphics[width=0.33\textwidth]{#1} & \includegraphics[width=0.33\textwidth]{#2} &  \includegraphics[width=0.33\textwidth]{#3} \\
(a) & (b) & (c) \\
\end{tabular}

\vspace*{0mm}\caption[]{#4
} \label{#5}

\vspace*{0mm}
\end{center}
\vspace*{0mm}
\end{figure*}}

\def\dcwfigurestart[#1,#2,#3,#4,#5,#6]{
\vspace*{-0.2in}\
\begin{center}
\begin{minipage}[c]{\columnwidth}{
\includegraphics[width=\columnwidth]{#1} 
\vspace*{-8mm}\caption[]{#2} \label{#3} \
}\end{minipage}\hspace*{\columnsep}\
\begin{minipage}[c]{\columnwidth}{
\includegraphics[width=\columnwidth]{#4} 
\vspace*{-8mm}\caption[]{#5}\label{#6} \
}\end{minipage}
}
\def\dcwfigureend[#1,#2,#3,#4,#5,#6]{
\vspace*{-0.15in}\
\begin{minipage}[c]{\columnwidth}{
\includegraphics[width=\columnwidth]{#1} 
\vspace*{-8mm}\caption[]{#2} \label{#3} \
}\end{minipage}\hspace*{\columnsep}\
\begin{minipage}[c]{\columnwidth}{
\includegraphics[width=\columnwidth]{#4} 
\vspace*{-8mm}\caption[]{#5}\label{#6} \
}\end{minipage}
\end{center}
\vspace*{-0.4in}\
}

\def\dcfigurerbiased[#1,#2,#3,#4,#5,#6]{
{
\begin{figure*}[h]
\vspace*{0in}\
\begin{center}
\begin{minipage}[c]{1.3\columnwidth}{
\includegraphics[width=\columnwidth]{#1} 
\vspace*{0mm}\caption[]{#2} \label{#3} \
}\end{minipage}\hspace*{\columnsep}\
\begin{minipage}[c]{0.7\columnwidth}{
\includegraphics[width=\columnwidth]{#4} 
\vspace*{0mm}\caption[]{#5}\label{#6} \
}\end{minipage}
\end{center}
\vspace*{0.0in}\
\end{figure*}
}
}

\def\dcfigurelbiased[#1,#2,#3,#4,#5,#6]{
{
\begin{figure*}
\vspace*{0in}\
\begin{center}
\begin{minipage}[c]{0.7\columnwidth}{
\includegraphics[width=\columnwidth]{#1} 
\vspace*{0mm}\caption[]{#2} \label{#3} \
}\end{minipage}\hspace*{\columnsep}\
\begin{minipage}[c]{1.2\columnwidth}{
\includegraphics[width=\columnwidth]{#4} 
\vspace*{0mm}\caption[]{#5}\label{#6} \
}\end{minipage}
\end{center}
\vspace*{-0.2in}\
\end{figure*}
}
}

\def\dcfigure[#1,#2,#3,#4,#5,#6]{
{
\begin{figure*}
\vspace*{0in}\
\begin{center}
\begin{minipage}[t]{\columnwidth}{
\includegraphics[width=\columnwidth]{#1} 
\vspace*{0mm}\caption[]{#2} \label{#3} \
}\end{minipage}\hspace*{\columnsep}\
\begin{minipage}[t]{\columnwidth}{
\includegraphics[width=\columnwidth]{#4} 
\vspace*{0mm}\caption[]{#5}\label{#6} \
}\end{minipage}
\end{center}
\vspace*{0mm}\
\end{figure*}
}
}

\def\dcfigureUnderAnotherFig[#1,#2,#3,#4,#5,#6]{
{
\begin{figure*}
\vspace*{0in}\
\begin{center}
\begin{minipage}[c]{\columnwidth}{
\includegraphics[width=\columnwidth]{#1} 
\vspace*{0mm}\caption[]{#2} \label{#3} \
}\end{minipage}\hspace*{\columnsep}\
\begin{minipage}[c]{\columnwidth}{
\includegraphics[width=\columnwidth]{#4} 
\vspace*{0mm}\caption[]{#5}\label{#6} \
}\end{minipage}
\end{center}
\vspace*{0in}\
\end{figure*}
}
}

\def\dcfiguretable[#1,#2,#3,#4,#5,#6]{
{
\begin{figure*}
\vspace*{-0in}\
\begin{center}
\begin{minipage}[c]{\columnwidth}{
\includegraphics[width=\columnwidth]{#1} 
\vspace*{-0mm}\caption[]{#2} \label{#3} \
}\end{minipage}\hspace*{\columnsep}\
\begin{minipage}[c]{\columnwidth}{
\vspace*{9mm}
\resizebox{\columnwidth}{!}{
\input{#4} 
}
\vspace*{0mm}\caption[]{#5}\label{#6} \
}\end{minipage}
\end{center}
\vspace*{-0.4in}\
\end{figure*}
}
}

\def\dcfigurebiasedtable[#1,#2,#3,#4,#5,#6]{
{
\begin{figure*}
\vspace*{-0.2in}\
\begin{center}
\begin{minipage}[c]{1.3\columnwidth}{
\includegraphics[width=\columnwidth]{#1} 
\vspace*{-0mm}\caption[]{#2} \label{#3} \
}\end{minipage}\hspace*{0.4\columnsep}\
\begin{minipage}[c]{0.75\columnwidth}{
\vspace*{0mm}
\resizebox{\columnwidth}{!}{
\input{#4} 
}
\caption[]{#5}\label{#6}
\vspace*{0mm} \
}\end{minipage}
\end{center}
\vspace*{-0.1in}\
\end{figure*}
}
}

\def\dssfigure[#1,#2,#3,#4,#5,#6]{
{
\begin{figure*}
\vspace*{0in}\
\begin{center}
\begin{minipage}[c]{4in}{
\includegraphics[width=4in]{#1}
\vspace*{0mm}\caption[]{#2} \label{#3} \
}\end{minipage}\hspace*{\columnsep}\
\begin{minipage}[c]{2in}{
\includegraphics[width=2in]{#4}
\vspace*{0mm}\caption[]{#5}\label{#6} \
}\end{minipage}
\end{center}
\vspace*{0in}\
\end{figure*}
}
}

\def\dsfigure[#1,#2,#3,#4,#5,#6]{
{
\begin{figure*}
\vspace*{0in}\
\begin{center}
\begin{minipage}[c]{3in}{
\includegraphics[width=3in]{#1}
\vspace*{0mm}\caption[]{#2} \label{#3} \
}\end{minipage}\hspace*{\columnsep}\
\begin{minipage}[c]{3in}{
\hspace*{\columnsep}\
\includegraphics[height=3in]{#4}
\vspace*{0mm}\caption[]{#5}\label{#6} \
}\end{minipage}
\end{center}
\vspace*{0in}\
\end{figure*}
}
}

\def\dsyfigure[#1,#2,#3,#4,#5,#6]{
{
\begin{figure*}
\vspace*{0in}\
\begin{center}
\begin{minipage}[c]{2.5in}{
\includegraphics[height=2.5in]{#1}
\vspace*{0mm}\caption[]{#2} \label{#3} \
}\end{minipage}\hspace*{0.5in}\
\begin{minipage}[c]{2.5in}{
\includegraphics[height=2.5in]{#4}
\vspace*{0mm}\caption[]{#5}\label{#6} \
}\end{minipage}
\end{center}
\vspace*{0in}\
\end{figure*}
}
}

\def\dyfigure[#1,#2,#3,#4,#5,#6]{
{
\begin{figure*}
\vspace*{0in}\
\begin{center}
\begin{minipage}[c]{3in}{
\includegraphics[height=3in]{#1} 
\vspace*{0mm}\caption[]{#2} \label{#3} \
}\end{minipage}\hspace*{0.5in}\
\begin{minipage}[c]{3in}{
\includegraphics[height=3in]{#4} 
\vspace*{0mm}\caption[]{#5}\label{#6} \
}\end{minipage}
\end{center}
\vspace*{0in}\
\end{figure*}
}
}

\def\dyoldfigure[#1,#2,#3,#4,#5,#6]{
{
\begin{figure*}
\vspace*{0.2in}\
\begin{center}
\begin{minipage}[c]{3in}{
\epsfysize=2.0in\
\hspace{0.5in}\
\epsfbox{#1}
\vspace*{-3mm}\caption[]{#2} \label{#3} \
}\end{minipage}\hspace*{0.25in}\
\begin{minipage}[c]{3in}{
\epsfysize=2.0in\
\hspace{0.5in}\
\epsfbox{#4}
\vspace*{-3mm}\caption[]{#5}\label{#6} \
}\end{minipage}
\end{center}
\vspace*{-0.4in}\
\end{figure*}
}
}

\def\cfiguredouble[#1,#2,#3,#4]{
\begin{figure}
\vspace*{0.2in}\
\begin{center}
\begin{minipage}[c]{1.5in}{
\epsfxsize=1.5in\
\epsfbox{#1}
}\end{minipage}\hspace*{0.1in}\
\begin{minipage}[c]{1.5in}{
\epsfxsize=1.5in\
\vspace{0.1in}\epsfbox{#2}
}\end{minipage}\vspace*{-0.10in} \caption[]{#3}\label{#4}
\end{center}
\vspace*{0in}\
\end{figure}
}

\def\wpfigure[#1,#2,#3,#4]{
\begin{figure*}
\vspace*{4mm}
\begin{center}

\includegraphics[width=#4]{#1} 

\vspace*{-3mm}\caption[]{#2
} \label{#3}

\vspace*{0mm}
\end{center}
\end{figure*}}

\def\wprfigure[#1,#2,#3,#4,#5]{
\begin{figure*}
\vspace*{4mm}
\begin{center}

\includegraphics[width=#4, angle=#5]{#1} 

\vspace*{-3mm}\caption[]{#2
} \label{#3}

\vspace*{-5mm}
\end{center}
\end{figure*}}

\def\DoubleFigureWSlide[#1,#2,#3,#4,#5,#6,#7,#8,#9]{
\begin{figure*}
\vspace*{#9}
\begin{center}
\begin{minipage}{#4}
\includegraphics[width=#4]{#1}
\vspace*{-3mm}\caption{#2
}\label{#3}
\end{minipage}
\hspace{2em}
\begin{minipage}{#8}
\includegraphics[width=#8]{#5}
\vspace*{-3mm}\caption{#6
}\label{#7}
\end{minipage}
\vspace*{-5mm}
\end{center}
\end{figure*}
}

\def\DoubleFigureW[#1,#2,#3,#4,#5,#6,#7,#8]{
\begin{figure*}
\vspace*{0in}
\begin{center}
\begin{minipage}{#4}
\includegraphics[width=#4]{#1}
\vspace*{-3mm}\caption{#2
}\label{#3}
\end{minipage}
\hspace{2em}
\begin{minipage}{#8}
\includegraphics[width=#8]{#5}
\vspace*{-3mm}\caption{#6
}\label{#7}
\end{minipage}
\vspace*{-5mm}
\end{center}
\end{figure*}
}

\def\DoubleFigureWHack[#1,#2,#3,#4,#5,#6,#7,#8]{
\begin{figure*}
\vspace*{0in}
\begin{center}
\begin{minipage}{3in}
\includegraphics[width=#4]{#1}
\vspace*{-3mm}\caption{#2
}\label{#3}
\end{minipage}
\hspace{2em}
\begin{minipage}{3in}
\includegraphics[width=#8]{#5}
\vspace*{-3mm}\caption{#6
}\label{#7}
\end{minipage}
\vspace*{-5mm}
\end{center}
\end{figure*}
}

\def\ddcfigure[#1,#2,#3,#4]{
\begin{figure*}
\vspace*{0.2in}\
\begin{center}
\begin{minipage}[c]{3in}{
\includegraphics[height=3in]{#1} 
}\end{minipage}\hspace*{0.5in}\
\begin{minipage}[c]{3in}{
\includegraphics[height=3in]{#2} 
}\end{minipage}\vspace*{-0.10in} \caption[]{#3}\label{#4}
\end{center}
\vspace*{-0in}\
\end{figure*}
}

\def\ddcfigureSlide[#1,#2,#3,#4,#5]{
\begin{figure*}
\vspace*{#5}\
\begin{center}
\begin{minipage}[c]{3in}{
\includegraphics[height=3in]{#1} 
}\end{minipage}\hspace*{0.5in}\
\begin{minipage}[c]{3in}{
\includegraphics[height=3in]{#2} 
}\end{minipage}\vspace*{-0.10in} \caption[]{#3}\label{#4}
\end{center}
\vspace*{-0.4in}\
\end{figure*}
}

\def\cxfigure[#1,#2,#3]{
\begin{figure}
\vspace*{4mm}
\begin{center}
 
\epsfxsize=2.5in\
\epsfbox{#1}\
 
\vspace*{-0.10in}\caption[]{#2
} \label{#3}
 
\vspace*{-5mm}
\end{center}
\vspace*{-2mm}
\end{figure}}

\usepackage{enumitem}
\setlist[description]{leftmargin=0.2cm}
\title{Accelerating Deep Convolutional Networks using low-precision and sparsity}
\name{Ganesh Venkatesh, Eriko Nurvitadhi, Debbie Marr}
\address{Intel Labs}
\begin{document}
%
\maketitle
\begin{abstract}
We explore techniques to significantly improve the compute efficiency and performance of Deep Convolution Networks without impacting their accuracy. To improve the compute efficiency, we focus on achieving high accuracy with extremely low-precision (2-bit) weight networks, and to accelerate the execution time, we aggressively skip operations on zero-values. We achieve the highest reported accuracy of \fixme{76.6\% Top-1/93\% Top-5} on the Imagenet object classification challenge with low-precision network\footnote{github release of the source code coming soon} while reducing the compute requirement by \fixme{$\sim$3$\times$} compared to a full-precision network that achieves similar accuracy. Furthermore, to fully exploit the benefits of our low-precision networks, we build a deep learning accelerator core, \dlc, that can achieve up to \fixme{1 TFLOP/$mm^2$} equivalent for single-precision floating-point operations (\fixme{$\sim$2 TFLOP/$mm^2$ for half-precision}), which is \fixme{$\sim$5$\times$} better than Linear Algebra Core~\cite{lac} and \fixme{$\sim$4$\times$} better than previous deep learning accelerator proposal~\cite{dadiannao}.
\end{abstract}

\begin{keywords}
Deep Neural Networks, Ternary-weight Convolutions, Accelerator 
\end{keywords}

\section{Introduction}
\label{sec:intro}

Deep Convolutional Neural Networks (\dnns) provide state-of-the-art accuracy for many computer vision and image analysis tasks~\cite{resnet}. The accuracy of \dnns\ is rapidly improving (for example, Top-5 error for Imagenet object classification challenge~\cite{imagenet} improved by \fixme{$>$6$\times$} in \fixme{5 years}) and is close to human-level accuracy in some cases~\cite{humanLevelAccuracy}. 

\dnns\ achieve higher accuracy by building more powerful models consisting of greater number of layers (network depth). However, this increase in network depth incurs a steep increase in compute and memory requirements. As a result, these networks are taking longer to train; multiple week training times are common even when using multiple GPU cards. Also, the greater compute requirements make \dnns\ harder to deploy, which has led to a lot of interest recently in specialized hardware solutions, both commercially~\cite{tpu, waveComputing} and in academia~\cite{eie, cnvlutin,dadiannao}.   

In this paper, we build on recently proposed low-precision convolution networks~\cite{xnorNet, ternaryConnect, ternaryWeightNetwork} to reduce compute requirements of these networks. While previous efforts compromise accuracy to gain compute efficiency, we aim to achieve similar (or slightly better) accuracy at a lower compute complexity. In particular, we train a low-precision variant of a 34-layer deep residual network (Resnet~\cite{resnet}) that attains higher accuracy than the \emph{vanilla} 18-layer deep resnet while requiring fewer floating-point operations (\fixme{$\sim$3$\times$} lower) and having a smaller model size (\fixme{7$\times$} smaller). Furthermore, to fully leverage the benefits of this low-precision network, we propose and evaluate a Deep Learning Accelerator Core, \dlc, that can achieve equivalent performance of up to \fixme{$\sim$1 Teraflop/$mm^2$} by skipping operations on zero values. We make the following contributions in this paper
\begin{description}
	\setlength{\itemsep}{2pt}
	\item[Demonstrate high accuracy using low-precision weights] Using low-precision 2-bit weight networks~\cite{ternaryWeightNetwork}, we achieve high accuracy of \fixme{76.6\%} Top-1/\fixme{93\%} Top-5 on Imagenet~\cite{imagenet}, \textit{\textbf{the highest reported with a low-precision network to our knowledge}} and within \fixme{1.3\%} of the 2015 Imagenet winner~\cite{resnet}. Furthermore, we show that in these low-precision networks, most of the floating-point operations operate on zero values -- both while training as well as inference. 
	\item[Define a deep-learning accelerator core,\dlc] We propose a deep-learning accelerator core, \dlc, that exploits the available sparsity in these networks to achieve high effective performance and can be applied to both training as well as inference.
	\item[Achieve high effective performance density of \fixme{$\sim$1 TFlop/$mm^2$}] Our evaluation, based on synthesis of our design in 14nm, shows that \dlc\ can sustain extremely high performance density, reaching up to \fixme{1 TFlop/$mm^2$} equivalent performance for many of the layers in current state-of-the-art Residual networks~\cite{resnet}. This is an order-of-magnitude higher performance density than the previously proposed deep learning accelerator~\cite{dadiannao}.
\end{description}
\section{Low-precision Deep Convolution Network}
\label{sec:dnn}

We first overview recent efforts on using low-precision weights as well as inducing sparsity in deep convolution networks, then motivate the need for our work. We then present how we train/fine-tune low-precision networks. We conclude this section with analysis on the sparsity available in these networks, quantifying the potential of acceleration by a hardware architecture that is optimized for efficient zero-skipping.

\subsection{Background: Lowering the precision of network weights}
\label{sec:ternaryWeightNetwork}

Several techniques have been proposed to lower the precision of network weights. These include approaches that reduce the weights to half-precision~\cite{baiduSpeechHalfPrecision}, binary value~\cite{xnorNet}, and ternary (2-bit) value~\cite{ternaryWeightNetwork}. While reducing the precision from float to half-precision can get almost \fixme{2$\times$} savings, the other approaches can achieve significantly higher savings (\fixme{16-32$\times$} smaller model, very few float multiplications). We use the approach from Lin et al.~\cite{ternaryWeightNetwork} that lowers the network weights to a ternary value with the options being \texttt{[-1, 0, 1]} as follows:
\begin{equation}
\label{eq:ternary}
W_{ter}(i,j,k,l) = 
	\begin{cases}
	 1 &: \textmd{ if W(i,j,k,l)} > w_{th} \\
    -1 &: \textmd{ if W(i,j,k,l)} < -1 * w_{th} \\
     0 &: \textmd{ otherwise}
     \end{cases}
\end{equation} 

While the previous proposals trade classification accuracy to achieve compute efficiency, we focus on low-precision networks that provide similar (or slightly better) accuracy than a full-precision network while requiring lesser compute. 

The recent DeepCompression~\cite{deepCompression} proposal looks at inducing sparsity in deep convolution networks via \emph{pruning} where certain weights/activations are clamped to zero value. Instead, we explore techniques that induce \emph{dynamic} sparsity in networks where the zero weights/activations can change across input samples and training phases. By doing so, the effective sparsity we achieve is much higher (\fixme{$>$2$\times$}) and we show its applicability on state-of-the-art networks.

\subsection{Evaluation}
\label{sec:evaluation}

We evaluate on two datasets - Cifar10 and Imagenet

\boldparagraph{Cifar10~\cite{cifar10}} This dataset consists of 60K small images from 10 different classes and we train on them using VGG~\cite{vgg} (baseline training recipe from~\cite{vggTorch}).

\boldparagraph{Imagenet~\cite{imagenet}} This dataset consists of 1.2 million images for training and 50,000 images for testing from 1000 classes. We train on this dataset using the Residual networks, Resnet~\cite{resnet} (baseline training recipe from~\cite{resnetTorch}).

We implement our proposals in Torch, a deep learning framework and our accuracy evaluations use single-crop testing. For evaluating our deep learning accelerator, we synthesize it in 14nm Intel process.

\subsection{Training Low-precision Networks}
\label{sec:ternaryWeightNetworkTrain}

We experiment with a number of different approaches to improve the accuracy and reduce the compute intensity of these extremely low-precision networks. In this section, we overview the different techniques -   

\begin{description}[leftmargin=0cm]
\setlength{\itemsep}{0pt}
\item[Pre-initialization from full-precision network] We train the network in full-precision for the first few iterations (15 iterations) and then switch over to the low-precision mode for rest of the training ($\sim$75 iterations). This improves the accuracy by almost \fixme{2\%} (Section~\ref{sec:imagenetAccResults}).  
\item[Skipping lowering of precision for parts of network] We do not lower the precision of the first layer in the network to minimize the information loss from the input image. This improves our Top-1 accuracy by \fixme{$\sim$0.5\%} (Section~\ref{sec:cifarAccResults}).
\item[Aggressive lowering of learning rate] We maintain a history of the train error and lower the learning rate when the train error does not go down for few iterations. This can improve our Top-1 accuracy by more than \fixme{1\%} in some instances. 
\item[Regularization] We regularize activations to reduce noise and induce more sparsity. As a side effect, this technique tends to smooth our convergence curve (Figure~\ref{fig:vggCifar10}). 
\item[ReLU threshold] We vary the thresholds on our rectifier units to induce higher sparsity and reduce noise.
\end{description}

\subsection{Sparsity in Low-precision Networks}
\label{sec:sparsity}

\cfigure[sparsityResnet34,{\figtitle{Sparsity in Low-precision Networks} Graph plots the fraction of operations  on non-zero operands when training Resnet-34 (34-layer deep) on Imagenet.},fig:sparsity]

In this section, we show that the vast majority of the operations in low-precision networks operate on zero values and hence, we can further reduce the compute requirements of these networks by skipping over operations on zero values (zero-skipping). The source of zero values in our network are as follows (i) \textbf{Rectifier Units (ReLU):} The rectifier units zero out activations below the threshold value (0.01), (ii) \textbf{Lowering the precision of weights:} The formula for lowering the precision effectively zeros out weights with small values (Eq.~\ref{eq:ternary}). 

The amount of sparsity is shown in Figure~\ref{fig:sparsity} for low-precision Resnet-34 training on the Imagenet dataset. The data shows that a small fraction (\fixme{16\%}) of operations during the forward pass (inference) operate on non-zero operands and only around \fixme{33\%} of operations during the backward pass operate on non-zero operands. This shows that we can improve performance by \fixme{3-6$\times$} by skipping operations on zero values. 

\section{Deep Learning Accelerator}
\label{sec:accl}

\cfigure[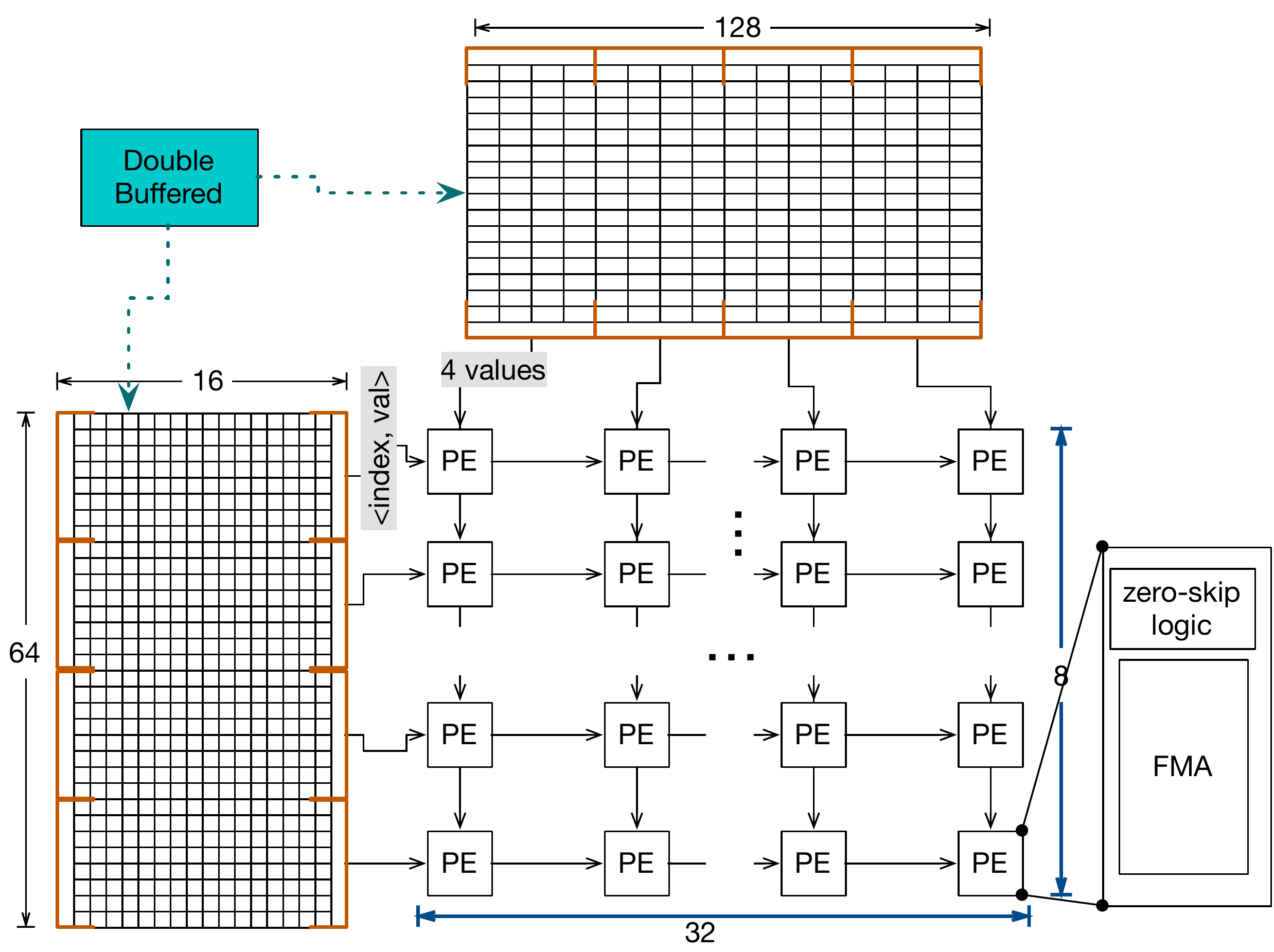, {\figtitle{\dlc\ Architecture} It is 2-D grid of processing elements where each one has floating-point units and logic to perform zero-skipping},fig:dlAccl]

\ignore{
\boldparagraph{Common operations in Deep Convolution Networks} The common operation in a deep convolution  network are:
\begin{description}
\setlength{\itemsep}{0pt}
\item[Convolution] A convolution operation operates on a pixel neighborhood (i.e. 3x3) and convolves them with a filter to get the new pixel values. A common technique for implementing this operation is to lower it to a matrix multiply~\cite{cudnn}.
\item[Matrix Multiply]  Matrix multiply is the formulation for multiple common operations such as fully-connected layers, 1x1 convolutions, linear classifiers as well as convolutions.
\item[Pointwise Operations] Deep convolution networks also employ many pointwise operations such as non-linearities (ReLU~\cite{relu}), batch normalization (inference), dropout etc.
\end{description}    
}

\dlc\ is a two-dimensional grid of processing elements with buffers for network weights and input feature map (Figure~\ref{fig:dlAccl}). The processing elements have arithmetic units, buffers for output feature map and control logic for skipping over operations on zero values to leverage the sparsity in low-precision networks. To enable skipping of zero operations, we assign multiple output buffers to each processing element (trading for greater number of buffers over few more arithmetic units) and when scheduling operations to update these output buffers, we skip the ones operating on zero values to accelerate the performance. 

Our accelerator supports the common operations in deep networks as follows:

\begin{description}
\setlength{\itemsep}{0pt}
\item[\texttt{mmOp:} Matrix Multiply] We use this to perform convolutions and the fully-connected layers.
\item[\texttt{ptWiseOp:} Pointwise Operation] The accelerators supports performing pointwise operations of the form \textmd{$<$op, val1, val2, val3$>$} on some or all the elements of the output buffer. The $op$ can be arithmetic operations (add/sub, mul-add), or a ternary control expression (?:). We use this to execute the non-linearity (Relu) and batch normalization (inference).
\end{description}

\boldparagraph{\dlc\ for training networks:} In training mode, we instantiate each \dlc\ with \fixme{512} single-precision floating-point units and all our datapaths are 32-bit wide. This allows \dlc\ to support the current standard training technique of using single-precision operations. Furthermore, our accelerator is able to sustain higher effective performance (\fixme{3-4 TeraOps/cycle}) by leveraging the \emph{dynamic} sparsity in these networks.

\boldparagraph{\dlc\ for inference:} In the inference mode, we instantiate each \dlc\ with \fixme{256} half-precision floating-point units, \fixme{256} half-precision adders and all the datapaths are 16-bit wide. This provides our accelerator with \fixme{$\sim$2$\times$} performance density boost. Similar to the training phase, the \dlc\ is able to sustain much higher throughput because of efficient zero-skipping. In addition to the low-precision networks, \dlc\ can also accelerate pruned networks~\cite{deepCompression}.

\cfigure[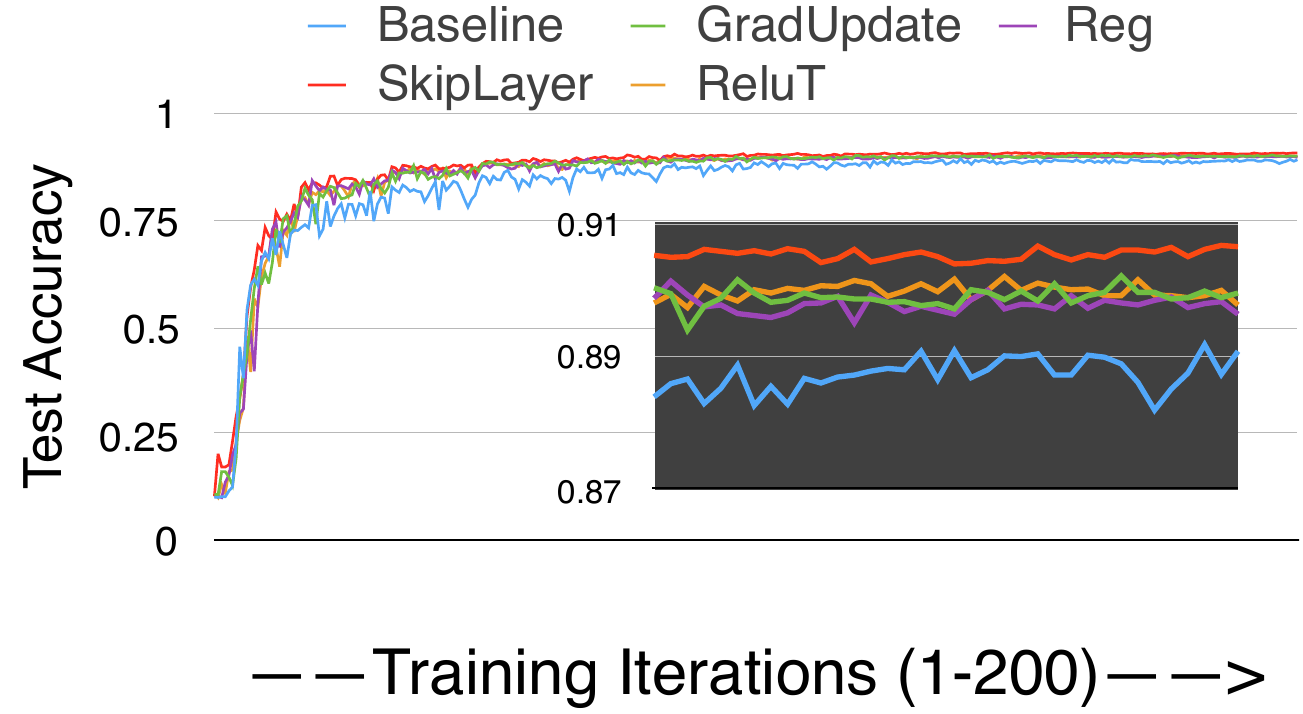,{\figtitle{VGG Convergence Graph for Cifar10} The dark graph inside is zoomed-in version of the final-few epochs of training.},fig:vggCifar10]

\section{Results}
\label{sec:results}

In this section, we evaluate the accuracy of our extremely low-precision networks as well as the performance they can attain on our deep learning accelerator, \dlc. 

\subsection{Accuracy of Low-precision Networks on Cifar10}
\label{sec:cifarAccResults}

We first evaluate low-precision variant of VGG network on Cifar10 dataset, shown in Figure~\ref{fig:vggCifar10}. Our baseline network replaces the regular convolution operation with its low-precision variant, which reduces the accuracy by \fixme{$\sim$3\%}. To improve the accuracy, we utilize the following techniques -- (i) \emph{Reg series}: we apply l1 regularization of activations which improves the accuracy by a small amount, (ii) \emph{GradUpdate series}:  to reduce overfitting, we backpropagate with zero error for samples that are classified correctly by the network and skip backpropagation step completely if the complete batch was correctly classified, and (iii) \emph{ReluT series}: we change the rectifier unit threshold (ReluT series) to \fixme{0.01} for the last few epochs of training to reduce the noise in activations, and (iv) \emph{SkipLayer series}: we compute the first convolution layer of the network in full-precision and convert the rest to low-precision variant. By doing so, we ensure that our model is able to capture all the information from the input image in the first layer while still benefiting from the efficiency of lower precision in the other layers of the network. The results show that these techniques improve the accuracy by \fixme{$\sim$1.6\%}. 

\subsection{Accuracy of Low-precision Networks on Imagenet}
\label{sec:imagenetAccResults}

In this section, we present the accuracy results on Imagenet dataset using multiple Resnet networks (2015 winner of Imagenet competition). We employ techniques that showed promise on Cifar10 dataset (regularization, ReLU threshold, skip precision lowering of first layer) and an additional trick of aggressively lowering the learning rate when the training accuracy stops improving. 

\begin{table}
\begin{tabular}{|c|c|c|c|c|}
\hline
Network & \multicolumn{2}{|c|}{Full-precision} & \multicolumn{2}{|c|}{2-bit precision}\\\cline{2-5}
Depth & Top-1~\cite{resnetTorch} & Top-5~\cite{resnetTorch} & Top-1 & Top-5 \\
\hline
Resnet-18 & 69.56 & 89.24 & - & - \\
Resnet-34 & 73.27 & 91.26 & 71.6 & 90.37 \\
Resnet-50 & 76 & 93 & 73.85 & 91.8 \\
Resnet-152 & 77.84 & 93.84 & 76.64 & 93.2 \\\hline
\end{tabular}
\caption{Accuracy of Resnet network on Imagenet dataset for different depths (first column suffix), regular full-precision, and the extremely low-precision 2-bit version. }
\label{tab:resnetAcc}
\end{table}

\boldparagraph{Lowering precision of trained networks} We start with the trained model of different Resnet networks, lower the precision of all the layers except for the first one, and train the resultant network. We report the accuracy numbers for the full-precision network and our low-precision 2-bit variant in Table~\ref{tab:resnetAcc}. The data shows that low-precision networks provide better accuracy as the network depth increases. Hence, just as their full-precision counterparts, the low-precision networks can also scale their depth to provide higher accuracy. Based on the data in Table~\ref{tab:resnetAcc}, we also observe that the low-precision variant of a larger network provide better accuracy than the regular full-precision network (low-precision Resnet-34/Resnet-50/Resnet-152 has higher accuracy than full-precision Resnet-18/Resnet-34/Resnet-50 respectively). This is an important result because the low-precision variant of a larger network needs less compute and has a smaller model size than the regular full-precision network. Hence, in effect, using the lower precision variant of a larger network achieves better accuracy and requires less compute than the original full-precision network. 

\cfigure[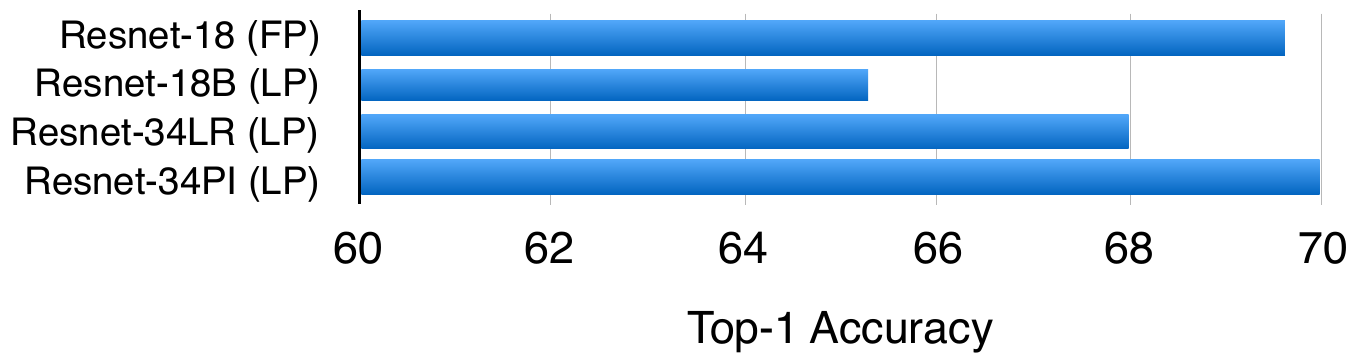,{\figtitle{Training Resnet-34 with low-precision weights} The graph shows the accuracy we obtain by training Resnet-34 using low-precision 2-bit weights by aggressively lowering the learning rate (\textit{Resnet-34LR}) as well as pre-initializing with a full-precision network(\textit{Resnet-34PI}). We obtain higher accuracy than previous work on a low-precision network~\cite{ternaryWeightNetwork} (Resnet-18B) as well as the full-precision Resnet-18 while requiring fewer computations than either.},fig:resnetAcc]

\boldparagraph{Training low-precision Resnet} We train Resnet with 34-layers and 2-bit weights. We try the following two techniques to improve accuracy (i) lowering the learning rate aggressively and (ii) training in full-precision for the first few iterations and switching over to low-precision after that. Our results are shown in Figure~\ref{fig:resnetAcc} -- we attain \fixme{$\sim$4.8\%} higher accuracy than previous work~\cite{ternaryWeightNetwork} using low-precision as well as slightly better accuracy than the regular full-precision variant of 18-layer Resnet. 

\subsection{Performance of \dlc\ on Low-precision Networks}
\label{sec:imagenetPerfResults}

\cfigure[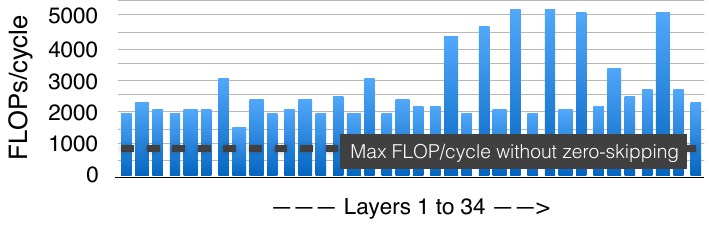,{\figtitle{Performance of \dlc\ on Resnet-34} The data shows the performance our accelerator can sustain for each layer in 34-layer deep Resnet. The graph shows that our accelerator gets significant performance boost (\fixme{1.8 - 5$\times$}) by skipping operations on zero-values and that our accelerator provides greater speed-up as we go deeper in the network because the layers get more and more sparse.},fig:resnetSpeedup]

Figure~\ref{fig:resnetSpeedup} shows the performance our accelerator can sustain for different convolution layers in Resnet-34. The accelerator can sustain up to \fixme{5K FLOP/cycle} (\fixme{2.78K FLOP/cycle on average}), which at \fixme{500MHz} translates to \fixme{2.5 Teraflops/second} (\fixme{1.34 Teraflops/second}). Furthermore, the graph shows that our accelerator provides better performance for the deeper layers of the network because of the greater sparsity in these layers. As a result, as we map deeper networks to our accelerator, we expect to get better performance because of its ability to exploit sparsity in the deeper layers. In single-precision mode, \dlc\ synthesizes to \fixme{2.2 $mm^2$} (\fixme{1.09 $mm^2$} in 16-bit mode) cell area in 14nm with pure ASIC flow for all the buffers and the arithmetic units (no optimized macro-blocks). So the \dlc\ compute IP has compute denstiy of \fixme{0.6 Teraflop/s/$mm^2$} and can exceed \fixme{1 Teraflop/s/$mm^2$} for deeper layers in the network.

\boldparagraph{Comparison to prior work} In comparison to DaDianNao Supercomputer~\cite{dadiannao}, one \dlc\ instance provides similar to slightly better performance than one node of DaDianNao while being \fixme{$>$4$\times$} smaller. Furthermore, our accelerator can provide higher performance as the sparsity in the network increases. In comparison to a recent work on zero-skipping~\cite{cnvlutin}, we achieve higher speed-ups by exploiting sparsity in both activations and weights. Unlike, EiE~\cite{eie}, we primarily focus on the convolution layers and support single-precision for the training phase. 

\ignore{
\boldparagraph{Towards a 50 TFLOP/sec deep learning inference engine} We can leverage the high performance of our \dlc\ accelerator to build a deep learning inference engine as follows: we can construct a 32-tile architecture where each tile consists of a general-purpose core (say 2-issue out-of-order RISC-V~\cite{boom}), \dlc\ accelerator and 256 KB cache. We run all the data-parallel operations on the \dlc\ accelerator and use general-purpose core for the control code and managing the accelerator. This would lead to a \fixme{$>$50 Teraflop} system within \fixme{$\sim$150 $mm^2$}. This is an important milestone because recent work in autonomous driving claims this is the performance that will enable a completely autonomous system~\cite{googleHotChips50TFlop}.}

\section{Conclusion}
\label{sec:conclude}

In this paper, we looked at improving the accuracy of extremely low-precision \dnns\ and encouraging greater dynamic sparsity in these low-precision networks to reduce their compute requirements. To fully leverage the efficiency of these low-precision networks, we developed and evaluated a deep learning accelerator, \dlc, that sustains high effective flops by skipping over operations on zero values. We demonstrate that these low-precision networks can attain high accuracy, achieving \fixme{76.6\% Top-1/93\% Top-5} accuracy on Imagenet and that the low-precision variant of a larger network can possibly achieve higher performance while requiring lesser compute than a regular full-precision network. Our \dlc\ evaluation shows that we can sustain up to \fixme{1 Teraflop/$mm^2$} equivalent which is a significant improvement over previous accelerator proposals.

\bibliographystyle{plain}
\bibliography{paper}

\end{document}